\documentclass[10pt,twocolumn,letterpaper]{article}

\usepackage{cvpr}
\usepackage{times}
\usepackage{epsfig}
\usepackage{graphicx}
\usepackage{amsmath}
\usepackage{amssymb}

\usepackage{mathtools}
\usepackage{breqn}
\usepackage{verbatim}
\makeatletter
\@namedef{ver@everyshi.sty}{}
\makeatother
\usepackage{pgfplots}
\usepackage{enumitem}
 \usepackage{array,multirow,graphicx}
 \usepackage{float}
\setlist[itemize]{noitemsep, nolistsep}

\usepackage[pagebackref=true,breaklinks=true,letterpaper=true,colorlinks,bookmarks=false]{hyperref}

\cvprfinalcopy 

\def\cvprPaperID{2330} 

\ifcvprfinal\fi
\begin{document}
\vspace{-5 mm}
\title{Looking for the Devil in the Details: Learning Trilinear Attention Sampling Network for Fine-grained Image Recognition}

\author{Heliang Zheng$^{1}\thanks{This work was performed when Heliang Zheng was visiting Microsoft Research as a research intern.}$  , Jianlong Fu$^{2}$, Zheng-Jun Zha$^{1}\thanks{Corresponding author.}$, Jiebo Luo$^{3}$ 
\\ \small $^1$University of Science and Technology of China, Hefei, China
\\ \small $^2$Microsoft Research, Beijing, China
\\ \small $^3$University of Rochester, Rochester, NY
\\ \small $^1$zhenghl@mail.ustc.edu.cn, zhazj@ustc.edu.cn, $^2$jianf@microsoft.com, $^3$jluo@cs.rochester.edu}

\maketitle
\thispagestyle{empty}
\vspace{-3 mm}
\begin{abstract}
Learning subtle yet discriminative features (e.g., beak and eyes for a bird) plays a significant role in fine-grained image recognition. Existing attention-based approaches localize and amplify significant parts to learn fine-grained details, which often suffer from a limited number of parts and heavy computational cost. In this paper, we propose to learn such fine-grained features from hundreds of part proposals by Trilinear Attention Sampling Network (TASN) in an efficient teacher-student manner. Specifically, TASN consists of 1) a trilinear attention module, which generates attention maps by modeling the inter-channel relationships, 2) an attention-based sampler which highlights attended parts with high resolution, and 3) a feature distiller, which distills part features into an object-level feature by weight sharing and feature preserving strategies. Extensive experiments verify that TASN yields the best performance under the same settings with the most competitive approaches, in iNaturalist-2017, CUB-Bird, and Stanford-Cars datasets. 



\end{abstract}

\vspace{-3 mm}
\section{Introduction}

Fine-grained visual categorization (FGVC) focuses on distinguishing subtle visual differences within a basic-level category (e.g., bird \cite{berg2014birdsnap,CUB-200-2011} and car \cite{StanfordCar,liu2016large,yang2015large}). Although the techniques of convolutional neural network (CNN) \cite{ResNet,Hinton_nips12,VGG19} for general image recognition \cite{krizhevsky2014cifar,ILSVRC15} have become increasingly practical, FGVC is still a challenging task where discriminative details are too subtle to be well-represented by traditional CNN. Thus the majority of efforts in the fine-grained community focuses on learning better representation for such subtle yet discriminative details.

\begin{figure}
\centering
\vspace{4 mm}
\includegraphics [width=0.48\textwidth]{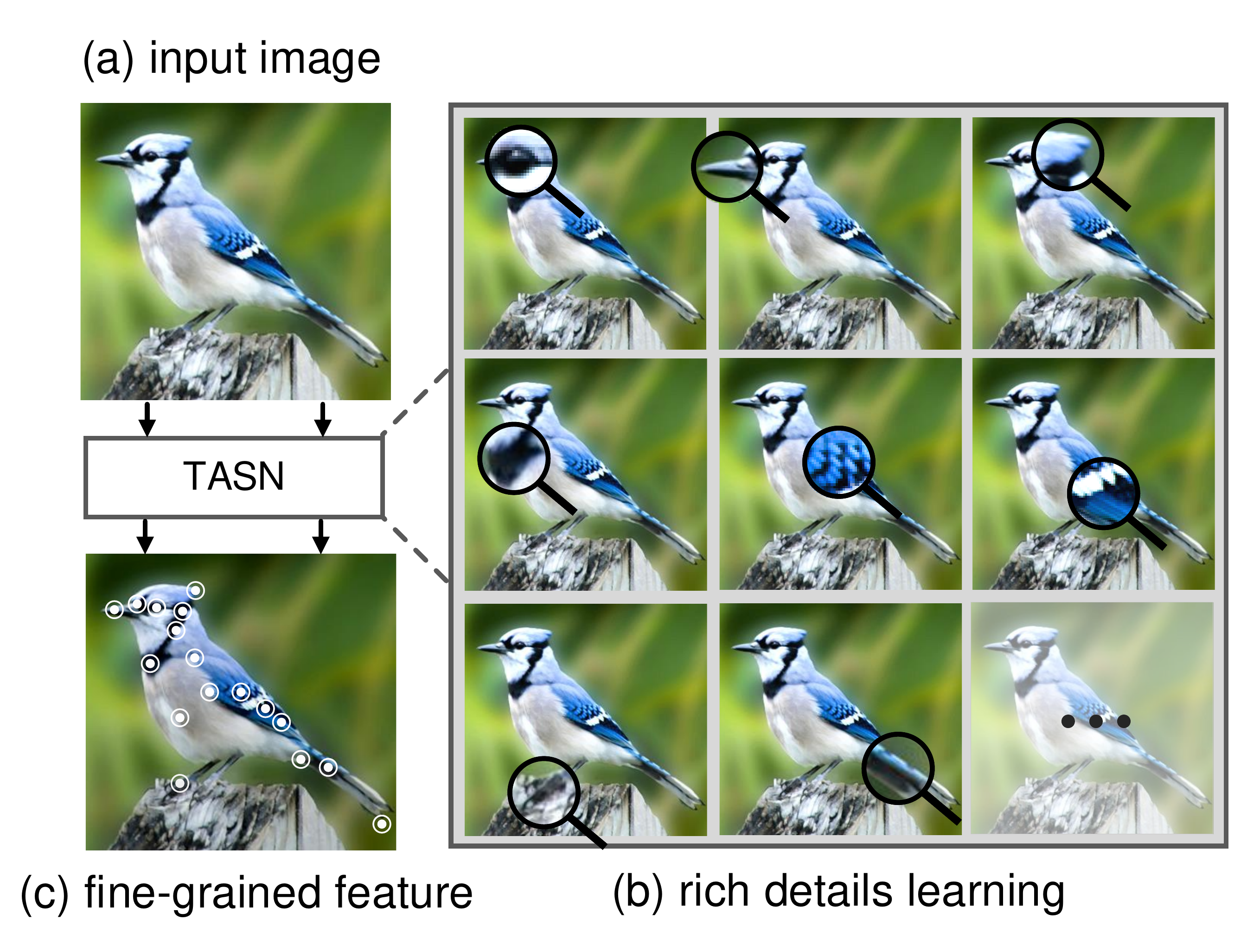}
\vspace{-4 mm}
\caption{An illustration of learning discriminative details by TASN for a ``bule jay.''  As shown in (b), TASN learns such subtle details by up-sampling each detail into high-resolution. And the white concentric circles in (c) indicates fine-grained details.}
\vspace{-3 mm}
\label{fig:fig1}
\end{figure}

Existing attention/part-based methods \cite{poseNorm,Fu_2017_CVPR,DBLP:journals/corr/WeiXW16,Zheng_2017_ICCV} try to solve this problem by learning part detectors, cropping and amplifying the attended parts, and concatenating part features for recognition. Although promising performance has been achieved, there are several critical issues in such a pipeline. Specifically, 1) the number of attention is limited and pre-defined, which restricts the effectiveness and flexibility of the model. 2) Without part annotations, it is difficult to learn multiple consistent (i.e., attending on the same part for each sample) attention maps. Although a well-designed initialization \cite{Fu_2017_CVPR,lam2017fine,Zheng_2017_ICCV} can benefit the model training, it is not robust and cannot handle the cases with uncommon poses. Moreover, 3) training CNNs for each part is not efficient. Such problems evolve as bottlenecks for the study on attention-based methods.

To address the above challenges, we propose a trilinear attention sampling network (TASN) which learns fine-grained details from hundreds of part proposals and efficiently distills the learned features into a single convolutional neural network. The proposed TASN consists of a trilinear attention module, an attention-based sampler, and a feature distiller.
First, the trilinear attention module takes as input feature maps and generates attention maps by self-trilinear product, which integrates feature channels with their relationship matrix. Since each channel of feature maps is transformed into an attention map, hundreds of part proposals can be extracted. 
Second, attention-based sampler takes as input an attention map as well as an image, and highlights attended parts with high resolution. Specifically, for each iteration, the attention-based sampler generates a detail-preserved image based on a randomly selected attention map, and a structure-preserved image based on an averaged attention map. The former learns fine-grained feature for a specific part, and the latter captures global structure and contains all the important details. 
Finally, A part-net and a master-net are further formulated as ``teacher'' and ``student,'' respectively. Part-net learns fine-grained features from the detail-preserved image and distills the learned features into master-net. And the master-net takes as input the structure-preserved image and refines a specific part (guided by the part-net) in each iteration. Such distillation is achieved by weight sharing and feature preserving strategies. Note that we adopt knowledge distilling introduced in \cite{hinton2014distilling} instead of concatenating part features, because the part number is large and not pre-defined.


Since the feature distiller transfers the knowledge from part-net into master-net via optimizing the parameters, 1) stochastic details optimization (i.e., randomly optimize one part in each iteration) can be achieved, which makes it practical to learn details from hundreds of part proposals, and 2) efficient inference can be obtained as we can use master-net to perform recognition in the testing stage. To the best of our knowledge, this work makes the first attempt to learn fine-grained features from hundreds of part proposals and represent such part features with a single convolutional neural network. Our contributions are summarized as follows:
\begin{itemize}
\item We propose a novel trilinear attention sampling network (TASN) to learn subtle feature representations from hundreds of part proposals for fine-grained image recognition.
\item We propose to optimize TASN in a teacher-student manner, in which fine-grained features can be distilled into a single master-net with high-efficiency.
\item We conduct extensive experiments on three challenging datasets (iNaturalist, CUB Birds and Stanford Cars), and demonstrate that TASN outperforms part-ensemble models even with a single stream.
\end{itemize}


The remainder of the paper is organized as follows. We describe related work in Section~\ref{rw}, and introduce our proposed TASN model in Section~\ref{method}. An evaluation on three widely-used datasets is presented in Section~\ref{exp}, followed by conclusions in Section~\ref{con}.
\begin{figure*}
\vspace{-6 mm}
\centering
\includegraphics [width=0.95\textwidth]{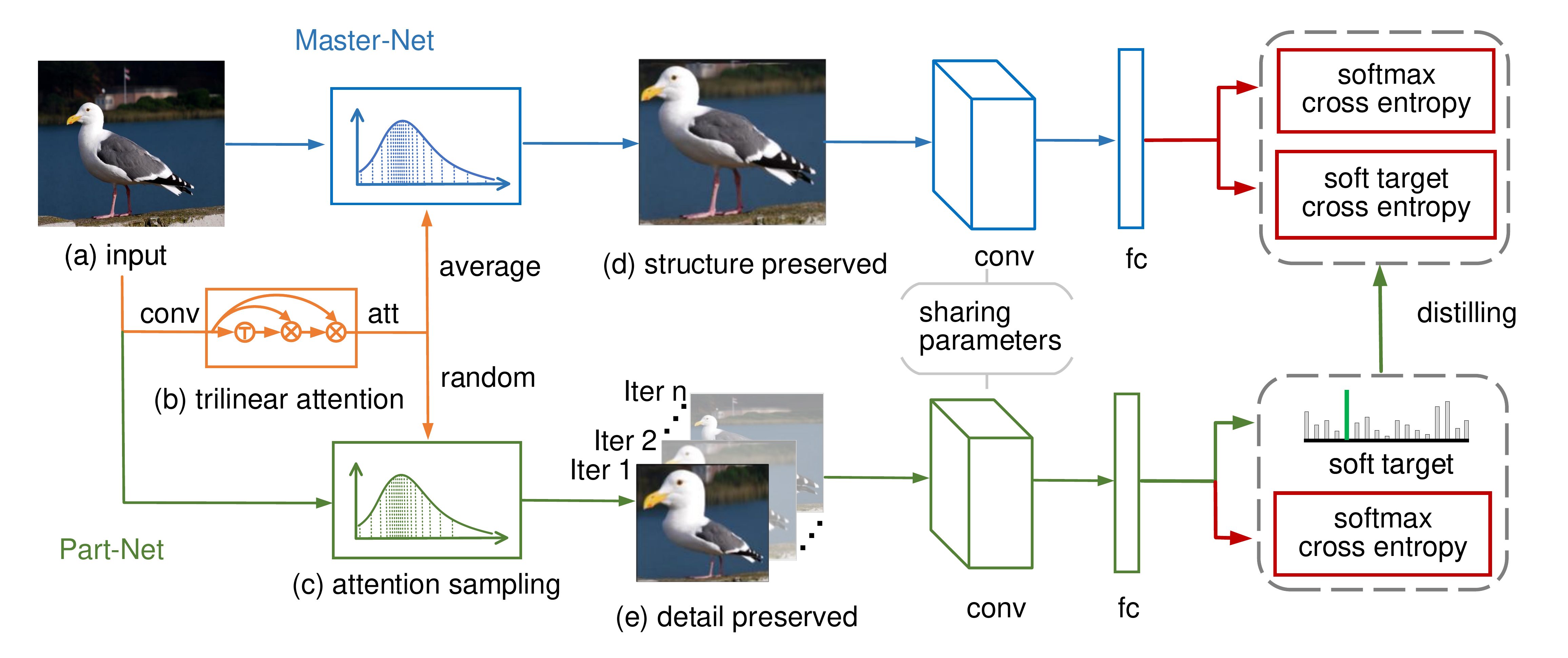}
\vspace{-2 mm}
\caption{Overview of the proposed Trilinear Attention Sampling Network (TASN). The trilinear attention module in (b) takes as input convolutional feature maps (denoted as ``conv''), and generates attention maps (denoted as ``att''). The attention sampling module in (c) further takes as input an attention map as well as the original image to obtain sampled images. Specifically, average pooling and random selection (in each iteration) are conducted over attention maps to obtain structure preserved image in (d) and detail preserved image in (e), respectively. The part-net (in green) learns fine-grained features from (e) and generates a soft target to distill such features into the master-net (in blue) via soft target cross entropy \cite{hinton2014distilling}. [Best viewed in color]}
\label{fig:archi}
\vspace{-4 mm}
\end{figure*}

\section{Related Works}
\label{rw}

\textbf{Attention Mechanism:}
As subtle yet discriminative details play an important role for Fine-Grained Image Recognition, learning to attend on discriminative parts is the most popular and promising direction. Thus various of attention mechanisms have been proposed in recent years \cite{Fu_2017_CVPR,li2017dynamic,sun2018multi,two-attention,Zheng_2017_ICCV}. DT-RAM \cite{li2017dynamic} proposed a dynamic computational time model for recurrent visual attention, which can attend on the most discriminative part in dynamic steps. RA-CNN \cite{Fu_2017_CVPR} proposed a recurrent attention convolutional neural network to recurrently learn attention maps in multiple (i.e., 3) scales. And MA-CNN \cite{Zheng_2017_ICCV} takes one step further to generate multiple (i.e., 4) consistency attention maps in a single scale by designing a channel grouping module. However, the attention numbers (i.e., 1, 3, 4, respectively) are pre-defined, which counts against the effectiveness and flexibility of the model.

Meanwhile, high-order attention methods are proposed in visual question answering (VQA) and video classification. Specifically, BAN \cite{kim2018bilinear} proposed a bilinear attention module to handle the relationship between image regions and the words in question, and Non-local \cite{wang2018non} calculates the dot production of features to represent the spatial and temporary relationship in video frames. Different from these works, our trilinear attention module conducts bilinear pooling to obtain the relationship among feature channels, which is further utilized to integrate such features to obtain third-order attention maps.

\textbf{Adaptive Image Sampling:}
To preserve fine-grained details for recognition, high input resolution ($448 \times 448$ v.s. $224 \times 224$) is widely adopted \cite{cui2018large,DBLP:journals/corr/WeiXW16,Zheng_2017_ICCV} and it can significantly improve the performance \cite{cui2018large}. However, high resolution brings large computational cost. More importantly, the importance of different regions are various, while directly zooming in images cannot promise different regions with different resolutions. STN \cite{SpatialTrans} proposed a non-uniformed sampling mechanism which performs well on MNIST datasets \cite{lecun1998gradient}. But without explicit guidance, it is hard to learn non-uniformed sampling parameters for sophisticated tasks such as fine-grained recognition, thus they finally learned two parts without non-uniformed sampling. SSN \cite{recasens2018learning} firstly proposed to use saliency maps as the guidance of non-uniformed sampling and obtained significant improvements. Different from them, our attention sampler 1) conduct non-uniformed sampling based on trilinear attention maps, and 2) decomposes attention maps into two dimensions to reduce spatial distortion effects.

\textbf{Knowledge Distilling:}
Knowledge distilling is firstly proposed by Hinton et al. \cite{hinton2014distilling} to transfer knowledge from an ensemble or from a large highly regularized model into a smaller, distilled model. The main idea is using soft targets (i.e., the predicted distribution of ensemble/large model) to optimize the small model, for it contains more information than the one-hot label. Such a simple yet effective idea inspires many researchers and has been further studied by \cite{DBLP:journals/corr/abs-1805-05532,yim2017gift}. In this paper, we adopt this technique to distill the learned details from part-net into master-net.

\section{Method}
\label{method}

In this section, we introduce the proposed Trilinear Attention Sampling Network (TASN), which is able to represent rich fine-grained features by a single convolutional neural network. TASN contains three modules, i.e., a trilinear attention module for details localization, an attention-based sampler for details extraction, and a feature distiller for details optimization.

An overview of the proposed TASN is shown in Figure~\ref{fig:archi}. Given an input image in (a), we first take it through several convolutional layers to extract feature maps, which is further transformed into attention maps by the trilinear attention module in (b). To learn fine-grained features for a specific part, we randomly select an attention map and conduct attention sampling over the input image using the selected attention map. The sampled image in (e) is named as detail-preserved image since it can preserve a specific detail with high resolution. Moreover, to capture global structure and contain all the important details, we average all the attention maps and again conduct attention sampling, such a sampled image in (d) is called structure-preserved image. We further formulate a part-net to learn fine-grained representation for detail-preserved images, and a master-net to learn the features for the structure-preserved image. Finally, the part-net generates soft targets to distill the fine-grained features into master-net via soft target cross entropy \cite{hinton2014distilling}.

\subsection{Details Localization by Trilinear Attention}
\label{sec:attention}
In this subsection, we introduce our trilinear attention module, which transfers convolutional feature maps into attention maps. As shown in previous work \cite{Part-discovery, deepFilter}, each channel of the convolutional features corresponds to a visual pattern, however, such feature maps cannot act as attention maps due to the lack of consistency and robustness \cite{wei2017selective, Zheng_2017_ICCV}. Inspired by \cite{ Zheng_2017_ICCV}, we transform feature maps into attention maps by integrating feature channels according to their spatial relationship. Note that such a process can be implemented in a trilinear formulation, thus we call it trilinear attention module.

Given an input image $\mathbf{I}$, we extract convolutional features by feeding it into multiple convolutional, batch normalization, ReLU, and pooling layers. Specifically, we use resnet-18 \cite{ResNet} as backbone. To obtain high-resolution feature maps for precise localization, we remove the two down-sampling processes from original resnet-18 by changing convolutional stride. Moreover, to improve the robustness of convolutional response, we increase the field of views \cite{chen2018deeplab} by appending two sets of dilated convolutional layers with multiple dilate rates. In the training stage, we added a softmax classifier to optimize such convolutional features.

\begin{figure}
\centering
\includegraphics [width=0.51\textwidth]{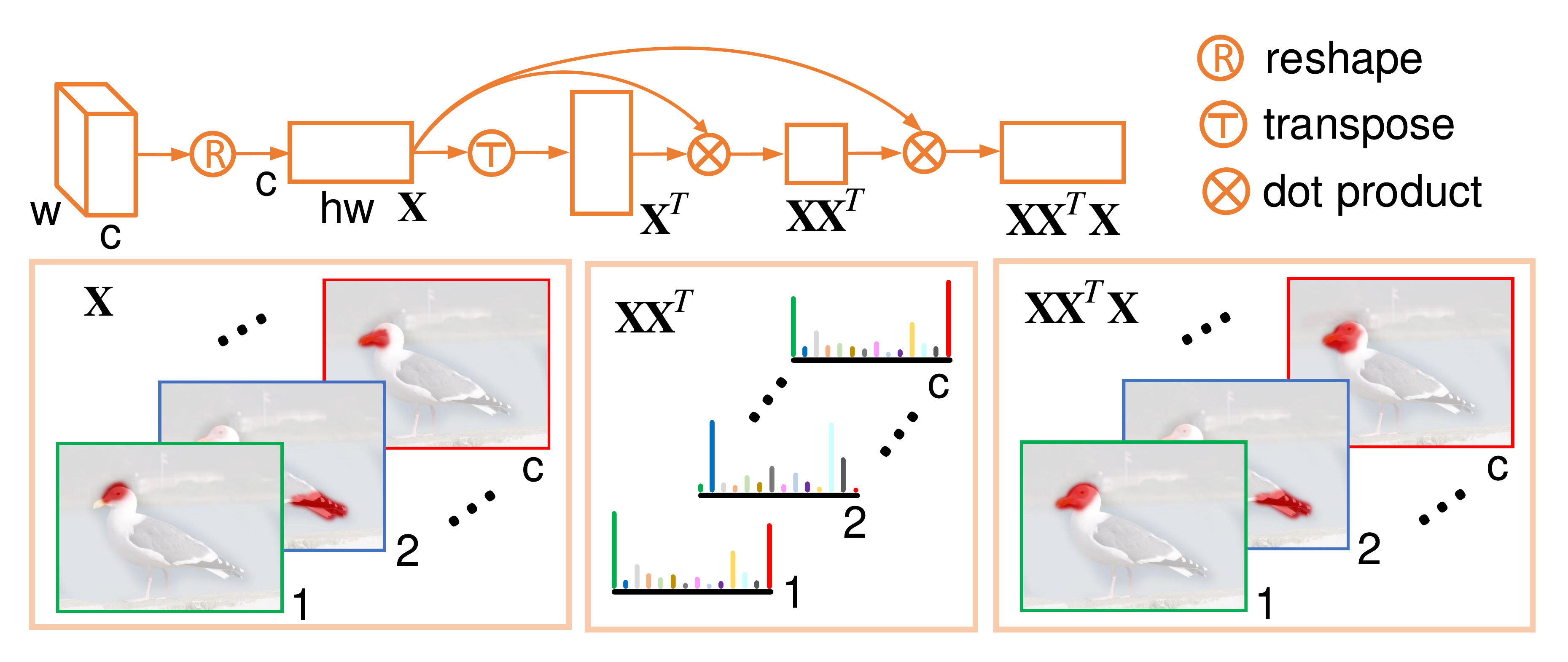}
\vspace{-6 mm}
\caption{An illustration the trilinear product. $\mathbf{X}$ indicates convolutional feature maps, and we can obtain inter-channel relationships by $\mathbf{X}\mathbf{X^T}$. After that, we integrate each feature map with its related ones to get trilinear attention maps via conducting dot production over $\mathbf{X}\mathbf{X^T}$ and $\mathbf{X}$.}
\vspace{-4 mm}
\label{fig:fig2}
\end{figure}

Assume that the feature maps is a tube with a dimension of $c\times h \times w$, where $c$, $h$ and $w$ indicate channel numbers, height, and width respectively. We reshape this feature into a matrix with a shape of $c\times hw$, which is denoted as $\mathbf{X} \in \mathbb{R}^{c\times hw}$. Then our trilinear function can be basically formulated as:
\begin{equation}\label{basic_trilinear}
\mathcal{M}_b(\mathbf{X}) \coloneqq (\mathbf{X}\mathbf{X^T})\mathbf{X},
\end{equation}
where $\mathbf{X}\mathbf{X^T}$ is the bilinear feature, which indicates the spatial relationship among channels. Specifically, $\mathbf{X}_i$ is the $i^{th}$ channel of feature maps, and $\mathbf{{XX^T}}_{i,j}$ indicates the spatial relationship between channel $i$ and channel $j$. To make feature maps more consistency and robust, we further integrate spatial relationship into feature maps by conducting dot production over $\mathbf{X}\mathbf{X^T}$ and $\mathbf{X}$, thus trilinear attention maps can be obtained (which is shown in Figure~\ref{fig:fig2}).

We further studied different normalization methods to improve the effectiveness of trilinear attention, and a detailed discussion can be found in Section~\ref{sec:exp_trilinear}. To the end, we adopt the following normalized trilinear attention:
\begin{equation}\label{trilinear}
\mathcal{M}(\mathbf{X}) \coloneqq \mathcal{N}(\mathcal{N}(\mathbf{X})\mathbf{X^T})\mathbf{X},
\end{equation}
where $\mathcal{N}(\cdot)$ indicates $softmax$ normalization over the second dimension of a matrix. Note that these two normalization functions have different meanings: The first one $\mathcal{N}(\mathbf{X})$ is spatial normalization which keeps each channel of feature maps within the same scale. And the second one is relationship normalization which is conducted over each relationship vector ${(\mathcal{N}(\mathbf{X})\mathbf{X^T})}_i$. We denote the output of the trilinear function in Equation~\ref{trilinear} as $\mathbf{M} \in \mathbb{R}^{c\times hw}$, i.e., $\mathbf{M} = \mathcal{M}(\mathbf{X})$. Finally, we reshape $\mathbf{M}$ into the shape of $c \times h \times w$, thus each channel of $\mathbf{M}$ indicates an attention map $\mathbf{M}_i \in \mathbb{R}^{h \times w}$.

\subsection{Details Extraction by Attention Sampling}
\label{sec:sampler}
In this subsection, we introduce our attention-based sampler, which takes as input an image as well as trilinear attention maps, and generates a structure-preserved image and a detail-preserved image.
The structure-preserved image captures the global structure and contains all the important details. Compared to the original image, the structure-preserved one removed the regions without fine-grained details, thus the discriminative parts can be better represented with high resolution. The detail-preserved image focuses on a single part, which can preserve more fine-grained details.

Given an image $\mathbf{I}$, we obtain structure-preserved image $\mathbf{I}_s$ and detail-preserved image $\mathbf{I}_d$ by conducting non-uniform sampling over different attention maps:
\begin{equation}\label{eqn:sampling}
\mathbf{I}_s = \mathcal{S}(\mathbf{I}, \mathcal{A}(\mathbf{M})), \quad
\mathbf{I}_d = \mathcal{S}(\mathbf{I}, \mathcal{R}(\mathbf{M})),
\end{equation}
where $\mathbf{M}$ is the attention maps, $\mathcal{S}(\cdot)$ indicates the non-uniform sampling function, $\mathcal{A}(\cdot)$ indicates average pooling over channels, and $\mathcal{R}(\cdot)$ indicates randomly selecting a channel from the input. We calculate the average of all attention maps to guide structure-preserved sampling, because such an attention map takes all the discriminative parts into consideration. And we randomly select one attention map for detail-preserved sampling, thus it can preserve the fine-grained details of this attended area with high resolution. With the training process going on, all attention maps have the opportunity to be selected, thus different fine-grained details can be asynchronously refined.

Our basic idea for attention-based sampling is considering the attention map as probability mass function, where the area with large attention value is more likely to be sampled. Inspired by the inverse-transform \cite{devroye1986sample}, we implement the sampling by calculating the inverse function of the distribution function. Moreover, we decompose attention maps into two dimensions to reduce spatial distortion effects.

Taking structure-preserved sampling for example, we first calculate the integral of the structure-preserved attention map $\mathcal{A}(\mathbf{M})$ over $x$ and $y$ axis:
\begin{equation}\label{eqn:distribution}
\begin{split}
&\mathcal{F}_x(n) \coloneqq  \sum_{j=1}^{n}{\max_{1 \le i \le w}{\mathcal{A}(\mathbf{M})_{i,j}}}, \\
&\mathcal{F}_y(n) \coloneqq  \sum_{i=1}^{n}{\max_{1 \le j \le h}{\mathcal{A}(\mathbf{M})_{i,j}}},
\end{split}
\end{equation}
where $w$ and $h$ are the width and height of the attention map, respectively. Note that we use $max(\cdot)$ function to decompose the attention map into two dimensions, because it is more robust than the alternative $sum(\cdot)$. We can further obtain the sampling function by:
\begin{equation}\label{eqn:trans}
\mathbf{\mathcal{S}(\mathbf{I}, \mathcal{A}(\mathbf{M}))}_{i,j} = \mathbf{I}_{\mathcal{F}^{-1}_x(i), \mathcal{F}^{-1}_y(j)}.
\end{equation}
where $\mathcal{F}^{-1}(\cdot)$ indicates the inverse function of $\mathcal{F}(\cdot)$. In a word, the attention map is used to calculate the mapping function between the coordinates of the original image and the sampled image.

\begin{figure}
\centering
\includegraphics [width=0.49\textwidth]{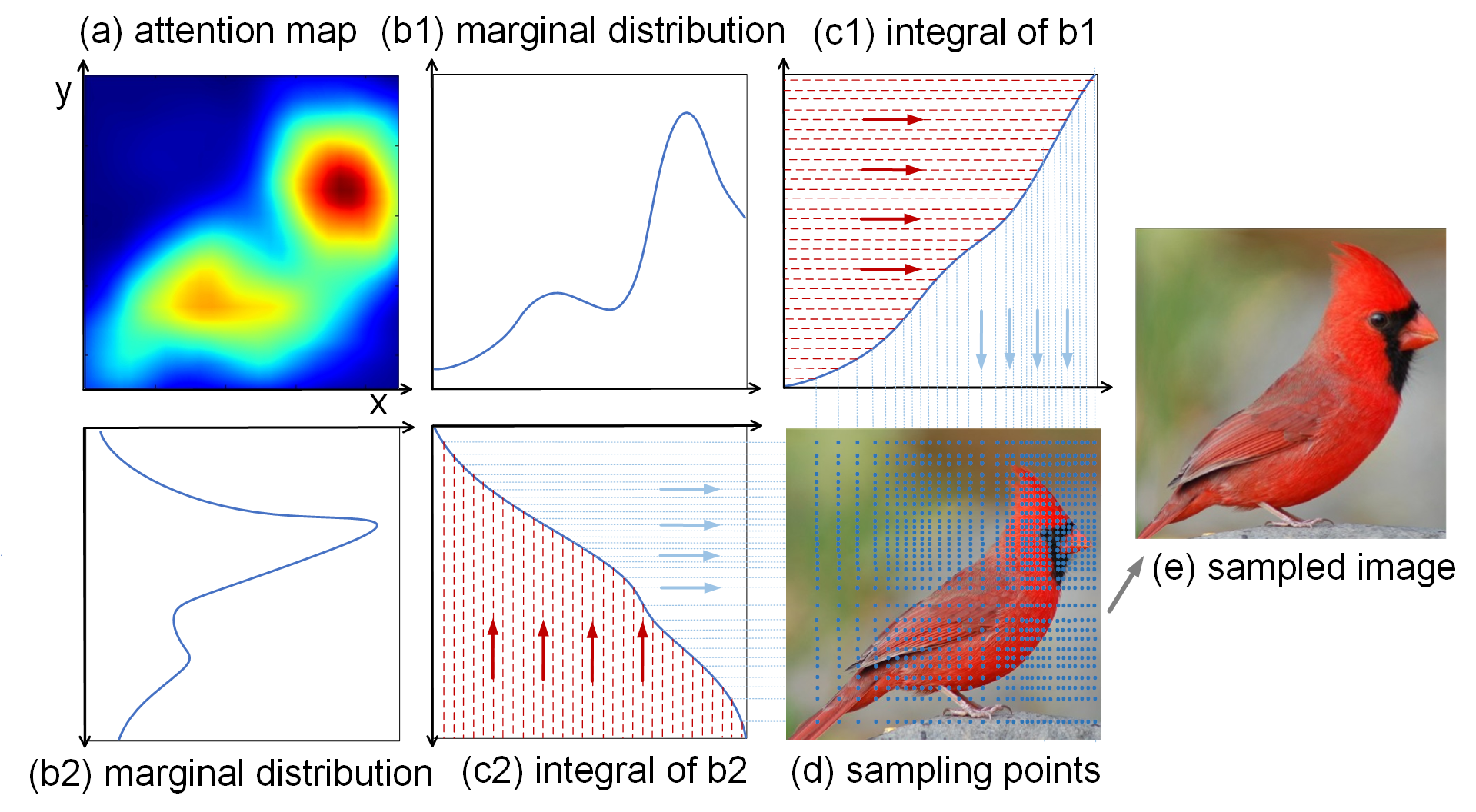}
\vspace{-4 mm}
\caption{An example of attention-based non-uniform sampling. (a) is an attention map with Gaussian distribution. (b1) and (b2) are the marginal distributions over $x$ and $y$ axis, respectively. (c1) and (c2) are the integrals of marginal distributions. (d) shows the sampling points by the blue dot, and (e) illustrates the sampled image. [Best viewed in color with zoom-in.]}
\vspace{-5 mm}
\label{fig:fig3}
\end{figure}

Such a sampling mechanism is illustrated in Figure ~\ref{fig:fig3}. Given an attention map in (a), we first decompose the map into two dimensions by calculating the max values over $x$ axis (b1) and $y$ axis (b2). Then the integral of (b1) and (b2) are obtained and shown in (c1) and (c2), respectively. We further calculate the inverse function of (c1) and (c2) in a digital manner, i.e., we uniformly sample points over the $y$ axis, and follow the red arrow (shown in (c1) and (c2)), and the blue arrow to obtain the values over $x$ axis. (d) shows the sampling points by blue dots, and we can observe that the regions with large attention values are allocated with more sampling points. Finally, (e) shows the result of the sampled image. Note that the example in Figure~\ref{fig:fig3} is a structure-preserved sampling case.

\subsection{Details Optimization by Knowledge Distilling}
\label{sec:distiller}
In this subsection, we introduce our details distiller, which takes as input a detail-preserved image and a structure-preserved image, and transfers the learned details from part-net to master-net in a teacher-student manner.

Specifically, for each iteration, the attention-based sampler introduced in Section ~\ref{sec:sampler} can provide a structure-preserved image (denoted as $\mathbf{I}_s$) and a detail-preserved one (denoted as $\mathbf{I}_d$). We first obtain the fully connected (fc) outputs by feeding these two images into the same backbone CNN (e.g., Resnet-50 \cite{ResNet}). The fc outputs are denoted as $\mathbf{z}_s$ and $\mathbf{z}_d$, respectively. Then the ``softmax'' classifier converts $z_s$ and $z_d$ into a probability vector $q_s$ and $q_d$, which indicates the predicted probability over each class. Taking $z_s$ for example:
\begin{equation}\label{eqn:prob}
q_s^{(i)} = \frac{exp(z_s^{(i)}/T)}{\sum_{j}{exp(z_s^{(j)}/T)}}, \\
\end{equation}
where $T$ is a parameter namely temperature, which is normally set to 1 for classification tasks. While in knowledge distilling, a large value for $T$ is important as it can produce a soft probability distribution over classes. We obtain the soft target cross entropy \cite{hinton2014distilling} for the master-net as:
\begin{equation}\label{loss_soft}
L_{soft}(q_s, q_d) = -  \sum_{i=1}^N{q_d^{(i)}log q_s^{(i)}} ,
\end{equation}
where $N$ denotes the class number. Finally, the objective function of the master-net can be drived by:
\begin{equation}\label{loss}
L(\mathbf{I}_s) = L_{cls}(q_s, y) + \lambda L_{soft}(q_s, q_d) ,
\end{equation}
where $L_{cls}$ represents the classification loss function, $y$ is a one hot vector which indicates the class label and $\lambda$ denotes loss weight of the two terms. The soft target cross entropy aims to distill the learned feature for fine-grained details and transfer such information to the master-net. As the attention-based sampler randomly select one part in each iteration, all the fine-grained details can be distilled to the master-net in training process. Note that the convolutional parameters are shared for part-net and master-net, which is important for distilling, while the sharing of fully connected layers is optional.

\section{Experiments}
\label{exp}
\subsection{Experiment setup}

\begin{table}
{\small
\caption{Detailed statistics of the three datasets used in this paper.}
\label{tab:dataset}
\begin{center}
\resizebox{0.99\columnwidth}{!}{
    \begin{tabular}{|c|c|c|c|}
    \hline
    Dataset  & \# Class & \# Train & \# Test\\
    \hline    \hline
    CUB-200-2011 \cite{CUB-200-2011} & 200 & 5,994 & 5,794\\
    Stanford-Car \cite{StanfordCar}  & 196 & 8,144 & 8,041\\
    iNaturalist-2017 \cite{van2018inaturalist}  & 5,089 & 579,184 & 95,986\\
    \hline
    \end{tabular}
    }
\end{center}
}
\vspace{-7 mm}
\end{table}

\textbf{Datasets:} To evaluate the effectiveness of our proposed TASN, we conducted experiments on three extensive and competitive datasets, namely Caltech-UCSD Birds (CUB-200-2011) \cite{CUB-200-2011},  Stanford Cars \cite{StanfordCar} and iNaturalist-2017\cite{van2018inaturalist}, respectively. The detailed statistics with category numbers and the standard training/testing splits can be found in Table~\ref{tab:dataset}. iNaturalist-2017 is the largest dataset for the fine-grained task. Compared with other datasets for this task, it contains 13 superclasses. Such a data distribution can provide a more convincing evaluation for the generalization ability of a model.

\textbf{Baselines:}
We compared our method to the following baselines due to their state-of-the-art performance and high relevance. Note that for a fair comparison, we did not include methods using 1) additional data (from the web or other datasets), 2) human-annotated part locations and 3) hierarchical labels (i.e., species, genus, and family). And all of the compared methods in each table share the same backbone unless specified otherwise.

\begin{itemize}
\item FCAN \cite{liu2016fully}: Fully convolutional attention network, which adaptively selects multiple attentions by reinforcement learning.
\item MDTP \cite{wang2016mining}: Mining discriminative triplets of patches, which utilize geometric constraints to improve the accuracy of patch localization.
\item DT-RAM \cite{li2017dynamic}: Dynamic computational time model for recurrent visual attention, which attends on the most discriminative parts by dynamic steps.
\item SSN \cite{recasens2018learning}: Saliency-based sampling networks, which conduct non-uniformed sampling based on saliency map in an end-to-end way.
\item MG-CNN \cite{multi-grained}: Multiple granularity descriptors, which leverage the hierarchical labels to generate comprehensive descriptors.
\item STN \cite{SpatialTrans}: Spatial transformer network, which conducts parameterized spatial transformation to obtain zoomed in or pose normalized objects.
\item RA-CNN \cite{Fu_2017_CVPR}: Recurrent attention CNN, which recurrently attends on discriminative parts in multi-scale.
\item MA-CNN \cite{Zheng_2017_ICCV}: Multiple attention CNN, which attends on multiple parts by their proposed channel grouping module in a weakly-supervised way.
\item MAMC \cite{sun2018multi}: Multi-attention multi-class constraint network, which learns multiple attentions by conducting multi-class constraint over attended features.
\item NTSN \cite{yang2018learning}: Navigator-Teacher-Scrutinizer Network, which is a novel self-supervision mechanism to effectively localize informative regions without the need of bounding-box/part annotations.
\item iSQRT-COV \cite{Li_2018_CVPR}: Towards faster training of global covariance pooling networks by iterative matrix square root normalization.

\end{itemize}
\begin{table}
{\small
\caption{Ablation experiments on attention module in terms of recognition accuracy on the CUB-200-2011 dataset.}
\label{tab:att}
\begin{center}
\resizebox{0.99\columnwidth}{!}{
    \begin{tabular}{|c|c|c|}
    \hline
    Attention & Description & Accuracy\\
    \hline    \hline
    \scalebox{0.9}{$\mathbf{X}$} & feature maps &   83.5   \\
    \scalebox{0.9}{$\mathbf{X}\mathbf{X^T}\mathbf{X}$} & trilinear attention &   84.9 \\
    \scalebox{0.9}{$\mathcal{N}(\mathbf{X})\mathbf{X^T}\mathbf{X}$} & spacial norm &  85.2 \\
    \scalebox{0.9}{$\mathcal{N}(\mathbf{X})\mathcal{N}(\mathbf{X})\mathbf{^T}\mathbf{X}$} & spacial norm &      84.3 \\
    \scalebox{0.9}{$\mathcal{N}(\mathbf{X}\mathbf{X^T}\mathbf{X})$} & spacial norm & 84.5 \\
    \scalebox{0.9}{$\mathcal{N}(\mathbf{X}\mathbf{X^T})\mathbf{X}$} & relation norm & 85.0  \\
    \scalebox{0.9}{$\mathcal{N}(\mathcal{N}(\mathbf{X})\mathbf{X^T})\mathbf{X}$} & spacial + relation &  \textbf{85.3}  \\
    \hline
    \end{tabular}
    }
\end{center}
}
 \vspace{-7 mm}
\end{table}

\textbf{Implementation:} We used open-sourced MXNet \cite{chen2015mxnet} as our code-base, and trained all the models on 8 Tesla P-100 GPUs. The backbones are are pre-trained on Imagenet \cite{ILSVRC15}, and all of the performances are single-crop testing results for a single model unless specially stated. We used SGD optimizer without momentum and weight decay, and the batch size was set to 96. The temperature in Equation~\ref{eqn:prob} is 10, and the loss weight $\lambda$ in Equation~\ref{loss} is 2. More implementation details can be referred to our code \url{https://github.com/researchmm/tasn}.
\label{details}

\subsection{Evaluation and analysis on CUB-200-2011}
\label{exp:CUB}

\textbf{Trilinear attention.}
\label{sec:exp_trilinear}
Table~\ref{tab:att} shows the impact of different normalization functions for the part-net in term of recognition accuracy. Specifically, we randomly select a channel of attention maps in each iteration in training stage, and conduct average pooling over attention maps for testing. All the models use Resnet-50 as the backbone with an input resolution of 224. It can be observed that trilinear attention maps can significantly outperform the original feature maps. Both the attention functions of $\mathcal{N}(\mathbf{X})\mathbf{X^T}\mathbf{X}$ and $\mathcal{N}(\mathbf{X}\mathbf{X^T})\mathbf{X}$ can improve the gain of trilinear attention. $\mathcal{N}(\mathbf{X})\mathcal{N}(\mathbf{X})\mathbf{^T}\mathbf{X}$ and $\mathcal{N}(\mathbf{X}\mathbf{X^T}\mathbf{X})$ bring a drop of performance, because such normalization functions is harmful for preserving spatial information. To this end, we adopt the last setting (of Table~\ref{tab:att}) in our TASN. Note that in the term $\mathcal{N}(\mathbf{X})\mathbf{X^T}$, $\mathcal{N}(\mathbf{X})$ indicates the region that a channel is focusing on and $\mathbf{X^T}$ denotes the feature of that region.

We further compared our trilinear attention module with ``self-attention'' \cite{vaswani2017attention}. Specifically, we followed \cite{vaswani2017attention} to obtain attention maps by $\mathbf{X^T}\mathbf{X}$, and the results show that the trilinear attention module can outperform self-attention module with 0.7\% points increases.

\begin{figure*}
\centering
\vspace{-4 mm}
\includegraphics [width=0.95\textwidth]{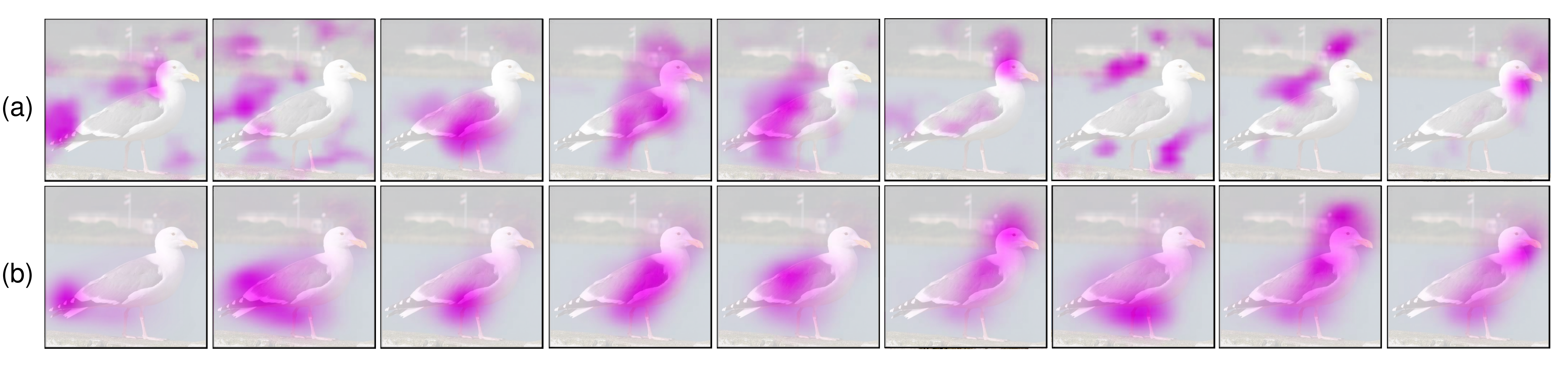}
\caption{A comparison of feature maps $\mathbf{X}$ in (a) and trilinear attention maps $\mathcal{N}(\mathcal{N}(\mathbf{X})\mathbf{X^T})\mathbf{X}$ in (b). Each column shows the same channel of feature maps and trilinear attention maps, and we randomly select nine channels for comparison. Compared to first-order feature maps, each channel of the trilinear attention maps focus on a specific part, without attending on background noises. [Best viewed in color]}
\label{fig:attentionmap}
\vspace{-4 mm}
\end{figure*}

\begin{table}
{\small
\caption{Ablation experiments on sampling module in term of classification accuracy on the CUB-200-2011 dataset.}
\label{tab:sam}
\begin{center}
\resizebox{0.83\columnwidth}{!}{
    \begin{tabular}{|c|c|c|}
    \hline
    Approach  & master-net & TASN\\
    \hline    \hline
    Resnet-50 \cite{ResNet} & 81.6 &  81.6   \\
    uniformed sampler & 84.1 &  85.8  \\
    sampler in SSN \cite{recasens2018learning} & 84.8 &  85.3  \\
    \hline
    our sampler &  \textbf{85.5} &  \textbf{87.0}   \\
    \hline
    \end{tabular}
    }
\end{center}
}
 \vspace{-3 mm}
\end{table}

\begin{table}

{\small
\caption{Ablation experiments on distilling module with different input resolutions.}
 \vspace{-1 mm}
 \label{tab:dis}
\begin{center}
\resizebox{0.9\columnwidth}{!}{
    \begin{tabular}{|c|c|c|c|c|}
    \hline
    Resolution  & 224 & 280 & 336 & 392\\
    \hline    \hline
    Resnet-50 \cite{ResNet} & 81.6 &  83.3 &85.0 & 85.6   \\
    master-net & 85.5 &  86.6 & 87.0 & 86.8 \\
    \hline
    TASN & \textbf{87.0} &  \textbf{87.3} & \textbf{87.9} & \textbf{87.9} \\
    \hline
    \end{tabular}
    }
\end{center}
}
 \vspace{-6 mm}
\end{table}

\textbf{Attention-based sampler.}
To demonstrate the effectiveness of our attention-based sampling mechanism, we compared our sampling mechanism with 1) uniformed sampling (by binarizing the attention maps) and 2) sampling operation introduced in SSN \cite{recasens2018learning}. We set the input attention maps to be same when comparing sampling mechanisms, and experiments were conducted on two cases, i.e., with and without part-net. All the models use Resnet-50 as the backbone and the input resolution is set to 224. The result in Table~\ref{tab:sam} shows that our sampling mechanism remarkably outperforms the baselines. SSN sampler obtains a better result than uniformed sampler without part-net, while the further improvements are limited when added part-net. These observations show that the spatial distortion caused by SSN sampler is harmful for preserving subtle details.

\textbf{Knowledge distilling.}
Table~\ref{tab:dis} reveals the impact of details distilling module with different input resolutions. We can observe consistency improvements by details distilling. The performance of Resnet-50 \cite{ResNet} is saturated with 85.6\%, and 448 input can not further improve the accuracy. Without distiller (i.e., master-net only), the performance is slightly dropped with 392 input (compared to 336 input), since it is difficult to optimize each detail with large feature resolutions (a similar drop can also be observed on Resnet-50 with 672 inputs).

Moreover, to study the attention selection strategy (i.e., ranking selection vs. random selection), we ranked attention maps by their response, and sample high response ones with large possibility, while the recognition performance dropped from 87.0\% to 86.8\%. The reason is that ranking makes some parts rarely picked, while such parts can also benefit details learning. We also conducted experiments on distilling two parts in each iteration, and the result is the same as distilling one part each time.

\textbf{Compared to sampling-based methods.}
We compare our TASN with three sampling-based methods: 1) uniformed sampling with high resolution (i.e., zoom in), 2) uniformed sampling with attention (i.e., crop) and 3) non-uniformed sampling proposed in SSN \cite{recasens2018learning}. As shown in Table~\ref{tab:sampling}, higher resolution can significantly improve fine-grained recognition performance by 4.9\% relatively. However, 448 input increases the computational cost (i.e., flops) by four times compared to 224 input. SSN \cite{recasens2018learning} obtains a better results than DT-RAM \cite{li2017dynamic}, and our TASN can further obtain 2.9\% relative improvement. Such improvements mainly come from two aspects: 1) a better sampling mechanism considering spatial distortion (1.2\%), and 2) a better fine-grained details optimizing strategy (1.7\%).

\begin{table}
{\small
\caption{Comparison with sampling-based methods in terms of classification accuracy on the CUB-200-2011 dataset.}
\label{tab:sampling}
\begin{center}
\resizebox{0.76\columnwidth}{!}{
    \begin{tabular}{|c|c|c|}
    \hline
    Approach  & Resolution & Accuracy\\
    \hline    \hline
    Resnet-50 \cite{ResNet}& 224 &  81.6   \\
    Resnet-50 \cite{ResNet}& 448 &  85.6   \\
    DT-RAM \cite{li2017dynamic}  & 224 &  82.8   \\
    SSN \cite{recasens2018learning}& 227  &  84.5   \\
        \hline
    TASN (ours) & 224  &  \textbf{87.0}   \\
    \hline
    \end{tabular}
    }
\end{center}
}
 \vspace{-7 mm}
\end{table}

\begin{table}
{\small
\caption{Comparison with part-based methods (all the results are reported in high-resolution setting) in terms of classification accuracy on the CUB-200-2011 dataset.}
\label{tab:part}
\begin{center}
\resizebox{0.85\columnwidth}{!}{
    \begin{tabular}{|c|c|c|}
    \hline
    Approach &  Backbone & Accuracy \\
    \hline \hline
    MG-CNN \cite{multi-grained} &  3$\times$VGG-16  & 81.7 \\
    ST-CNN \cite{SpatialTrans} &  3$\times$Inception-v2  & 84.1 \\
    RA-CNN \cite{Fu_2017_CVPR} &  3$\times$VGG-19  &  85.3   \\
    MA-CNN \cite{Zheng_2017_ICCV} & 3$\times$VGG-19  &  85.4   \\
    \hline
    TASN (ours) & 1$\times$VGG-19 &  \textbf{86.1}  \\
    TASN (ours) & 3$\times$VGG-19 &   \textbf{87.1}  \\
    \hline
    \hline
    MAMC \cite{sun2018multi} & 1$\times$Resnet-50 &  86.5   \\
    NTSN \cite{yang2018learning} & 3$\times$Resnet-50 &  87.3   \\
    \hline
    TASN (ours) & 1$\times$Resnet-50 &  \textbf{87.9}   \\
    \hline
    \end{tabular}
    }
\end{center}
}
 \vspace{-7 mm}
\end{table}

\textbf{Compared to attention-based part methods.}
In Table~\ref{tab:part}, we compare our TASN to attention-based parts methods. For a fair comparison, 1) high-resolution input is adopted by all methods and 2) the same backbone numbers are used. It can be observed that for VGG based methods, our TASN outperforms all the baselines even with only one backbone. Moreover, after ensembling three backbones (trained with different parameter settings), TASN can improve the performance by 1.9\% over the best 3 parts model MA-CNN \cite{Zheng_2017_ICCV}. Moreover, our 3 streams result can also outperform 6 streams MA-CNN (86.5\%) with a margin of 0.7\%. We do not ensemble more streams as the model ensemble is beyond this work. For Resnet-50 based method: compared with the state-of-the-art single-stream MAMC \cite{sun2018multi}, our TASN achieves a remarkable improvement by 1.6\%. Moreover, although NTSN \cite{yang2018learning} ($K = 2$) concatenates global feature with two part features, our single-stream TASN still can achieve 0.6\% points increases.

\textbf{Combining with second-order feature learning methods.}
In Table~\ref{tab:bilinear}, we exhibit that \textit{our TASN} learns a strong first-order representation, which \textit{can further improve the performance of second-order feature methods}. Specifically, compared to the best second-order methods iSQRT-COV \cite{Li_2018_CVPR}, our TASN 2k first-order feature outperforms their 8k feature with an improvement by 0.7\%, which shows the effectiveness of our TASN. Moreover, we transfer their released code to our framework and obtain an accuracy of 89.1\%, which shows the compatibility of these two methods. Note that for a fair comparison, we follow their settings and predict the label of a test image by averaging prediction scores of the image and its horizontal flip.

\begin{table}
{\small
\caption{Extensive experiments on combining second-order feature learning methods.}
\label{tab:bilinear}
\begin{center}
\resizebox{0.85\columnwidth}{!}{
    \begin{tabular}{|c|c|c|}
    \hline
    Approach & Dimension & Accuracy\\
    \hline \hline
    iSQRT-COV \cite{Li_2018_CVPR} & 8k & 87.3 \\
    iSQRT-COV \cite{Li_2018_CVPR} &  32k & 88.1 \\
    \hline
    TASN (ours) &  2k &  87.9   \\
    TASN + iSQRT-COV & 32k &  \textbf{89.1}   \\
    \hline
    \end{tabular}
    }
\end{center}
}
 \vspace{-5 mm}
\end{table}

\subsection{Evaluation and analysis on Stanford-Car}
\label{exp:car}
\begin{table}
{\small
\caption{Component analysis in terms of classification accuracy on the Stanford-Car dataset.}

\label{tab:car_self}
\begin{center}
\resizebox{0.83\columnwidth}{!}{
    \begin{tabular}{|c|c|c|c|c|}
    \hline
    Approach & Backbone & Accuracy \\
    \hline \hline
    Baseline & 1$\times$VGG-19 &  88.6   \\
    master-net & 1$\times$VGG-19 &  90.3   \\
    \hline
    TASN & 1$\times$VGG-19 &  92.4   \\
    TASN (ensemble) & 2$\times$VGG-19 &  93.1   \\
    TASN (ensemble)  & 3$\times$VGG-19 &  \textbf{93.2}   \\
    \hline
    \end{tabular}
    }
\end{center}
}
 \vspace{-4 mm}
\end{table}

\begin{table}
{\small
\caption{Comparison in terms of classification accuracy on the Stanford-Car dataset.}
\label{tab:car}
\begin{center}
\resizebox{0.83\columnwidth}{!}{
    \begin{tabular}{|c|c|c|c|c|}
    \hline
    Approach & Backbone & Accuracy \\
    \hline \hline
    FCAN \cite{liu2016fully} &  3$\times$VGG-16  & 91.3 \\
    MDTP \cite{wang2016mining} &  3$\times$VGG-16  & 92.5 \\
    RA-CNN \cite{Fu_2017_CVPR} &  3$\times$VGG-19  &  92.5   \\
    MA-CNN \cite{Zheng_2017_ICCV} & 3$\times$VGG-19  &  92.6   \\
    \hline
    TASN (ours) & 1$\times$VGG-19 &  92.4 \\
    TASN (ours) & 3$\times$VGG-19 &  \textbf{93.2}   \\
    \hline
    \hline
    MAMC \cite{sun2018multi} & 1$\times$Resnet-50 &  92.8   \\
    NTSN \cite{yang2018learning} & 3$\times$Resnet-50 &  93.7   \\
    \hline
    TASN (ours) & 1$\times$Resnet-50 & \textbf{93.8}  \\
    \hline
    \end{tabular}
    }
\end{center}
}
 \vspace{-5 mm}
\end{table}


Table~\ref{tab:car_self} shows the result of VGG-19 baseline, our master-net, a single TASN model, and TASN ensemble results. We can observe 1.9\% relative improvements by structure preserved sampling and further improvements of 2.3\% by the full model. Table~\ref{tab:car} compares TASN with attention-based parts methods. Specifically, TASN with single VGG-19 achieves comparable results with 3 streams part methods. And our ensembled 3 streams TASN outperforms the best 3 streams part learning methods MA-CNN \cite{Zheng_2017_ICCV}. Compared to their 5 streams result (92.8\%), our result is still better. For Resnet-50 based method, we compare our TASN to the state-of-the-art method MAMC \cite{sun2018multi}, and achieve 1.1\% improvements. Moreover, our single-stream TASN can achieve slightly better performance than NTSN \cite{yang2018learning}, which concatenates a global feature with two part features.

\subsection{Evaluation and analysis on iNaturalist 2017}
\label{exp:inat}
\begin{table}
{\small
\caption{Comparison in terms of classification accuracy on the iNaturalist 2017 dataset.}
 \vspace{-3 mm}
\label{tab:inat}
\begin{center}
\resizebox{1.0\columnwidth}{!}{
    \begin{tabular}{|c|c|c|c|c|}
    \hline
    Super Class & \# Class & Resnet         \cite{ResNet} & SSN \cite{recasens2018learning} & TASN \\
    \hline \hline
    Plantae & 2101 & 60.3 &  63.9 & \textbf{66.6} \\
    Insecta & 1021 & 69.1 &  74.7 & \textbf{77.6} \\
    Aves & 964 & 59.1 &  68.2 & \textbf{72.0} \\
    Reptilia & 289 & 37.4 &  43.9 & \textbf{46.4} \\
    Mammalia & 186 & 50.2 &  55.3 & \textbf{57.7} \\
    Fungi & 121 & 62.5 &  64.2 & \textbf{70.3} \\
    Amphibia & 115 & 41.8 &  50.2 & \textbf{51.6} \\
    Mollusca & 93 & 56.9 &  61.5 & \textbf{64.7} \\
    Animalia & 77 & 64.8 &  67.8 & \textbf{71.0} \\
    Arachnida & 56 & 64.8 &  73.8 & \textbf{75.1} \\
    Actinopterygii &53 & 57.0 &  60.3 & \textbf{65.5} \\
    Chromista & 9 & 57.6 &  57.6 & \textbf{62.5} \\
    Protozoa & 4 & 78.1 &  79.5 & \textbf{79.5} \\
    \hline
    Total & 5089 & 59.6 &  65.2 & \textbf{68.2} \\
    \hline
    \end{tabular}
    }
\end{center}
}
 \vspace{-9 mm}
\end{table}

We also conduct our TASN on the largest fine-grained dataset, i.e., iNaturalist 2017. We compare to Resnet \cite{ResNet} baseline and the best sampling method SSN \cite{recasens2018learning}. All the models use Resnet-101 as the backbone with an input resolution of 224. As there are 13 superclasses in this dataset, we re-implement SSN \cite{recasens2018learning} with their released code to obtain the performance on each superclass. The results are shown in Table~\ref{tab:inat}, and we can observe that TASN outperforms Resnet baseline and SSN on every superclass. It is notable that compared to Resnet-101, TASN significantly improves the performance, especially on Reptilia (improved by 24.0\%, relatively) and Aves (improved by 21.8\%, relatively), which indicates that such superclasses contain more fine-grained details.
\vspace{-2 mm}
\section{Conclusion}
\label{con}
In this paper, we proposed a trilinear attention sampling network for fine-grained image recognition, which can learn rich feature representations from hundreds of part proposals. Instead of ensembling multiple part CNNs, we adopted knowledge distilling method to integrate fine-grained features into a single stream, which is not only efficient but also effective. Extensive experiments in CUB-Bird, iNaturalist 2017 and Stanford-Car demonstrate that TASN is able to outperform part-ensemble models even with a single stream. In the future, we will further study the proposed TASN in the following directions: 1) attention selection strategy, i.e., learning to select which details should be learned and distilled instead of randomly selecting, 2) conduct attention-based sampling over convolutional features instead of only over images, and 3) extend our work to other vision tasks, e.g., object detection and segmentation.

\textbf{Acknowledgement:} This work was supported by the National Key R\&D Program of China under Grant 2017YFB1300201, the National Natural Science Foundation of China (NSFC) under Grants 61622211 and 61620106009 as well as the Fundamental Research Funds for the Central Universities under Grant WK2100100030.

{\small
\bibliographystyle{ieee}
\bibliography{egbib}
}

\end{document}